\documentclass[conference]{IEEEtran}
\IEEEoverridecommandlockouts
\usepackage{cite}
\usepackage{amsmath,amssymb,amsfonts}
\usepackage{textcomp}
\usepackage{xcolor}
\def\BibTeX{{\rm B\kern-.05em{\sc i\kern-.025em b}\kern-.08em
    T\kern-.1667em\lower.7ex\hbox{E}\kern-.125emX}}

\usepackage{xspace}
\usepackage{booktabs}
\usepackage[ruled,vlined,linesnumbered]{algorithm2e}

\usepackage{url}

\usepackage{graphicx}
\graphicspath{{.}{./Image/}}
\usepackage{subfigure}

\usepackage{wrapfig}

\newcommand{\shortname}{FedTune\xspace}

\begin{document}

\title{\shortname: Automatic Tuning of Federated Learning Hyper-Parameters from System Perspective}

\author{
Huanle Zhang$^1$, 
Mi Zhang$^{2,3}$, 
Xin Liu$^1$, 
Prasant Mohapatra$^1$, 
and Michael DeLucia$^4$ \\
$^1$Department of Computer Science, University of California, Davis, USA\\
$^2$The Ohio State University, USA  $^3$Michigan State University, USA\\
$^4$DEVCOM Army Research Laboratory, USA \\

\{dtczhang, xinliu, 
pmohapatra\}@ucdavis.edu,
mizhang.1@osu.edu, 
michael.j.delucia2.civ@army.mil
\vspace{-0.3in}
}


\maketitle

\begin{abstract}

In Federated Learning (FL), hyper-parameters significantly affect the training overhead in terms of computation time, transmission time, computation load, and transmission load.
The current practice of manually selecting FL hyper-parameters puts a high burden on FL practitioners since various applications have different training preferences. In this paper, we propose \shortname, an automatic hyper-parameter tuning algorithm tailored to applications' diverse system requirements in FL training. \shortname is lightweight and flexible, achieving 8.48\%-26.75\% improvement for different datasets compared to using fixed FL hyper-parameters.



\end{abstract}


\section{Introduction}

Federated learning (FL) has been applied to a wide range of applications such as mobile keyboard~\cite{gboard} and speech recognition~\cite{appleSpeech} on top of mobile devices~\cite{fedrolex2022neurips} and Internet of Things (IoT)~\cite{dledge2020bookchapter,fliotvision2022ieeeiotm}.
Compared to other model training paradigms~(e.g., centralized machine learning~\cite{machine_learning}, conventional distributed machine learning~\cite{distributed_learning}), FL has unique properties such as massively distributed,
significant unbalanced, and non-IID data distribution~\cite{FedAvg17aistats}. 
%
In addition to the common hyper-parameters of model training such as learning rates, optimizers, and mini-batch sizes, FL has unique hyper-parameters, including aggregation algorithms and participant selection~\cite{Oort21OSDI,pyramidfl2022mobicom}.
Fortunately, these FL hyper-parameters do not affect the FL convergence property. Many FL algorithms such as FedAvg~\cite{FedAvg17aistats}, have been proved to converge to the global optimum under different FL hyper-parameters~\cite{fedAvgConverge20iclr}\cite{fedNova20neurips}. However,  they can significantly affect the training overhead of reaching the final model.

In this paper, we focus on the training overhead.
Specifically, computation time (CompT), transmission time (TransT), computation load (CompL), and transmission load (TransL) are the four most important system overhead. CompT measures how long an FL system spends in model training; TransT represents how long an FL system spends in model parameter transmission between the clients and the server; CompL is the number of Floating-Point Operation~(FLOP) that an FL system consumes; and TransL is the total data size transmitted between the clients and the server. 

Application scenarios can have different training preferences in terms of CompT, TransT, CompL, and TransL. Consider the following examples: (1) attack and anomaly detection in computer networks~\cite{anomaly21ajrcos}
is time-sensitive (CompT and TransT) as it needs to adapt to malicious traffic rapidly; (2) smart home control systems for indoor environment automation~\cite{smart_home21jaihc}, e.g., heating, ventilation, and air conditioning~(HVAC), are sensitive to computation (CompT and CompL) because sensor devices are limited in computation capabilities; (3) traffic monitoring systems for vehicles~\cite{traffic20access}
are communication-sensitive (TransT and TransL) because cellular communications are usually adopted to provide city-scale connectivity. 

A few papers have studied FL training performance under different hyper-parameters~\cite{flField}. However, they do not consider CompT, TransT, CompL, and TransL together, which are essential from the system's perspective. In addition, it is challenging to tune multiple hyper-parameters in order to achieve diverse training preferences, especially when we need to optimize multiple system aspects. For example, it is unclear how to select hyper-parameters to build an FL training solution that is both CompT and TransL-efficient.  

\textbf{Contributions.}
This paper targets a new research problem of optimizing the hyper-parameters for FL from the system perspective. To do so, we formulate the system overhead in FL training and conduct extensive measurements to understand FL training performance. 
To avoid manual hyper-parameter selection, we propose \shortname, an algorithm that automatically tunes FL hyper-parameters during model training, respecting application training preferences. Our evaluation results show that \shortname achieves a promising performance in reducing the system overhead. 


\section{Related Work}

Hyper-Parameter Optimization (HPO) is a field that has been extensively studied~\cite{hpo_survey}. Many classical HPO algorithms, e.g., Bayesian optimization~\cite{bayesianOptimization}, successive halving~\cite{halfSuccessing}, and hyperband~\cite{hyperband}, are designed to optimize hyper-parameters of machine learning models. 

Designing HPO methods for FL, however, is a new research area. Only a few works have touched FL HPO problems. For example, FedEx is a general framework to optimize the round-to-accuracy of FL by exploiting the Neural Architecture Search (NAS) techniques of weight-sharing, which improves the baseline by several percentage points~\cite{fedex21neurips}; FLoRA determines the global hyper-parameters by selecting the hyper-parameters that have good performances in local clients~\cite{flora}. 
However, existing works cannot be directly applied to our scenario of optimizing FL hyper-parameter for different FL training preferences for two reasons. First, CompT (in seconds), TransL (in seconds), CompL~(in FLOPs), and TransL (in bytes) are not comparable with each other. Incorporating training preferences in HPO is not trivial. Second, hyper-parameter tuning needs to be done during the FL training. No ``comeback'' is allowed as the FL model keeps training until its final model accuracy. Otherwise, it will cause significantly more system overhead.


\section{Understanding the Problem}

We first quantify the system overheads of FedAvg to illustrate the problem.  
FedAvg minimizes the following objective 

\begin{equation}
    f(w) = \sum_{k=1}^{K} \frac{n_k}{n} F_k(w) \quad \text{where} \quad F_k(w) = \frac{1}{n_k} \sum_{i \in \mathcal{P}_k} f_i(w)
\end{equation}
where $f_i(w)$ is the loss of the model on data point $(x_i, y_i)$, that is,  $f_i (w) =\ell (x_i, y_i; w)$,  $K$ is the total number of clients, $\mathcal{P}_k$ is the set of indexes of data points on client $k$, with $n_k = |\mathcal{P}_k|$, and $n$ is the total number of data points from all clients, i.e., $n = \sum_{k=1}^K n_k$. Due to the large number of clients in a typical FL application (e.g., millions of clients in the Google Gboard project~\cite{gboard}), a common practice is to randomly select a small fraction of clients in each training round. In the rest of this paper, we refer to the selected clients as participants and denote $M$ as the number of participants on each training round. 
Each participant makes $E$ training passes over its local data in each round before uploading its model parameters to the server for aggregation. Afterward, participants wait to receive an updated global model from the server, and a new training round starts.

\subsection{System Model}


Assume that clients are homogeneous in terms of hardware~(e.g., CPU/GPU) and network~(e.g., transmission speeds). Let $b_{k, r}$ indicates whether client $k$ participates at the training round $r$. Then, we have $\sum_{k=1}^{K} b_{k, r} = M$, i.e., each round selects $M$ participants. The number of training rounds to reach the final model accuracy is denoted by $R$, which is unknown \textit{a priori} and varies when different sets of FL hyper-parameters are used in FL training. 
CompT, TransT, CompL, and TransL can be formulated as follows. 

\noindent
\textbf{Computation Time (CompT)}. If client $k$ is selected on a training round, it spends time in local training. The local training delay can be represented by $C_1 \cdot E \cdot n_k$, where $C_1$ is a constant. It is proportional to its number of data points~(i.e., $n_k$) because $n_k$ decides the number of local updates (number of mini-batches) for one epoch, and each local update includes one forward-pass and one backward-pass. 
The computation time of the training round $r$ is determined by the slowest participant and thus is represented by $ C_1 \cdot E \cdot \max_{k=1}^K b_{k, r} \cdot n_k$. In total, the computation time of an FL training can be formulated as 
\begin{equation}
    CompT = C_1 \cdot E \cdot \sum_{r=1}^{R}\max_{k=1}^K b_{k, r} \cdot n_k  
    \label{equ:compT}
\end{equation}

\noindent
\textbf{Transmission Time (TransT)}. Each participant on a training round needs one download and one upload of model parameters from and to the server~\cite{flField}. Thus, the transmission time is the same for all participants on any training round, i.e., a constant $C_2$. The total transmission time is represented by 
\begin{equation}
    TransT = C_2 \cdot R  
    \label{equ:transT}
\end{equation}


\noindent
\textbf{Computation Load (CompL)}.
Client $k$ causes $C_3 \cdot E \cdot n_k$ computation load if it is selected on a training round, where $C_3$ is a constant. The computation load of the training round $r$ is the summation of each participant's computation load and thus is $C_3 \cdot E \cdot \sum_{k=1}^K b_{k, r} \cdot  n_k$. We can formulate the overall computation load as
\begin{equation}
    CompL = C_3 \cdot E \cdot \sum_{r=1}^R \sum_{k=1}^K b_{k, r} \cdot  n_k
    \label{equ:compL}
\end{equation}

\noindent
\textbf{Transmission Load (TransL)}. Since each training round selects $M$ participants, the transmission load for a training round is $C_4 \cdot M$ where $C_4$ is a constant. The total number of training rounds is $R$, and thus, the total transmission load of an FL training is represented by 
\begin{equation}
    TransL = C_4 \cdot R \cdot M
    \label{equ:transL}
\end{equation} 


In the experiments, we assign the model's number of FLOPs for one input to $C_1$ and $C_3$, and the model's number of parameters to $C_2$ and $C_4$.

\subsection{Measurement Study}
\label{sect:measurement}


We conduct measurements to study the system overhead when different FL hyper-parameters are used for training. We use the Google speech-to-command dataset~\cite{speechToCommandDataset}.
Please refer to Section \ref{sect:eva_overall} for the training setup. 
The speech-to-command dataset meets the data properties of FL: massively distributed, unbalanced, and non-IID.
The measurement study investigates the FL training overhead in terms of the following three hyper-parameters.

\begin{itemize}
    \item The number of participants (i.e., $M$). It is well-known that more participants on each training round have a better round-to-accuracy performance~\cite{FedAvg17aistats}. 
    In the measurement study, we set $M$ to 1, 10, 20, and 50. 
    
    \item The number of training passes (i.e., $E$). Increasing the number of training passes as a method to improve communication efficiency has been adopted in several works, such as FedAvg~\cite{FedAvg17aistats} and FedNova~\cite{fedAvgConverge20iclr}. 
    In the measurement study, we set $E$ to 0.5, 1, 2, 4, 8, where 0.5 means that only half of each client's local data are used for local training in each round. 
    
    \item Model complexity. We also investigate how the model complexity influences the training overhead if a target accuracy is met. 
    We use ResNet~\cite{resnet} to build different models, as listed in Table \ref{tab:model}.
\end{itemize}

\begin{table}[!t]
    \centering
\scalebox{0.95}{
    \begin{tabular}{c c c c c}
        Model & ResNet-10 & ResNet-18 & ResNet-26 & ResNet-34 \\
        \toprule
        \#BasicBlock & [1, 1, 1, 1] & [2, 2, 2, 2] & [3, 3, 3, 3] & [3, 4, 6, 3] \\
        \#FLOP ($\times 10^6$) & 12.5 & 26.8 & 41.1 & 60.1 \\
        \#Params ($\times 10^3$) & 79.7 & 177.2 & 274.6 & 515.6 \\
        Accuracy & 0.88 & 0.90 & 0.90 & 0.92 \\
        \bottomrule
    \end{tabular}
    }
    \vspace{0.05in}
    \caption{Different models used for the measurement study.}
    \label{tab:model}
\end{table}

\begin{figure}[!t]
\centering
\centerline{
\subfigure[Computation Time]{
    \includegraphics[width=1.7in]{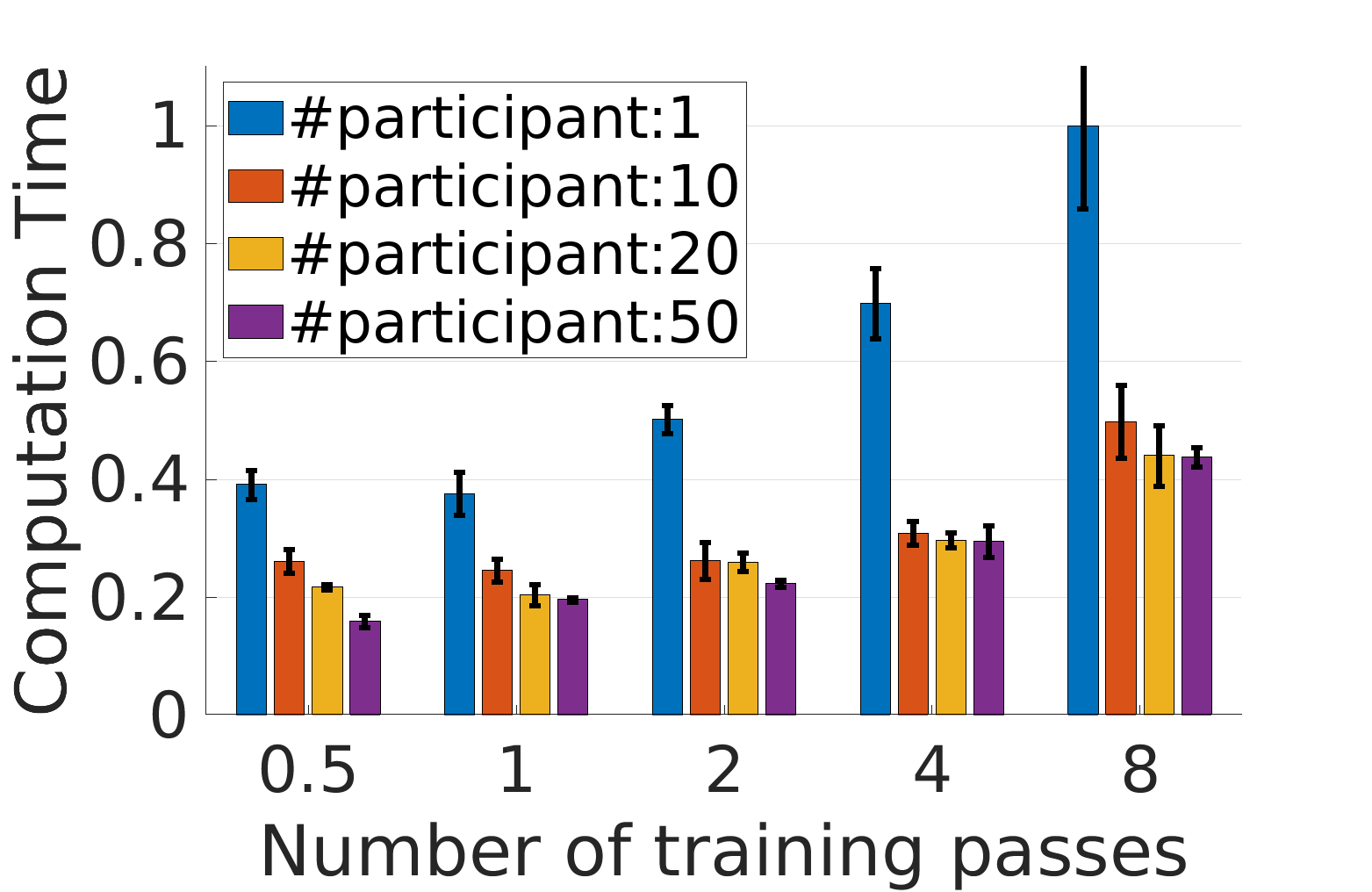}
    \label{fig:measure_M_E_compT}
}
\subfigure[Transmission Time]{
    \includegraphics[width=1.7in]{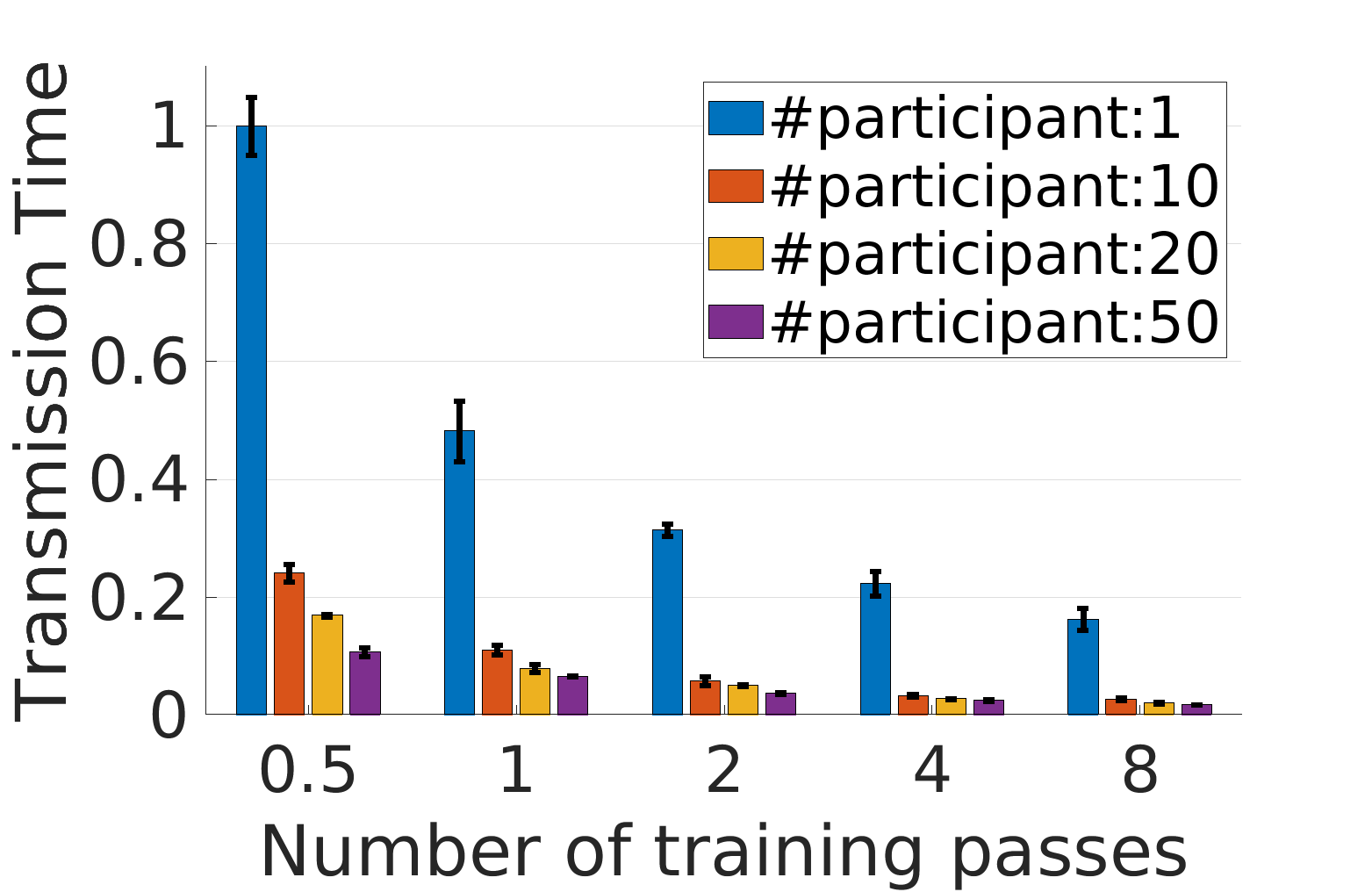}
    \label{fig:measure_M_E_transT}
}
}
\centerline{
\subfigure[Computation Load]{
    \includegraphics[width=1.7in]{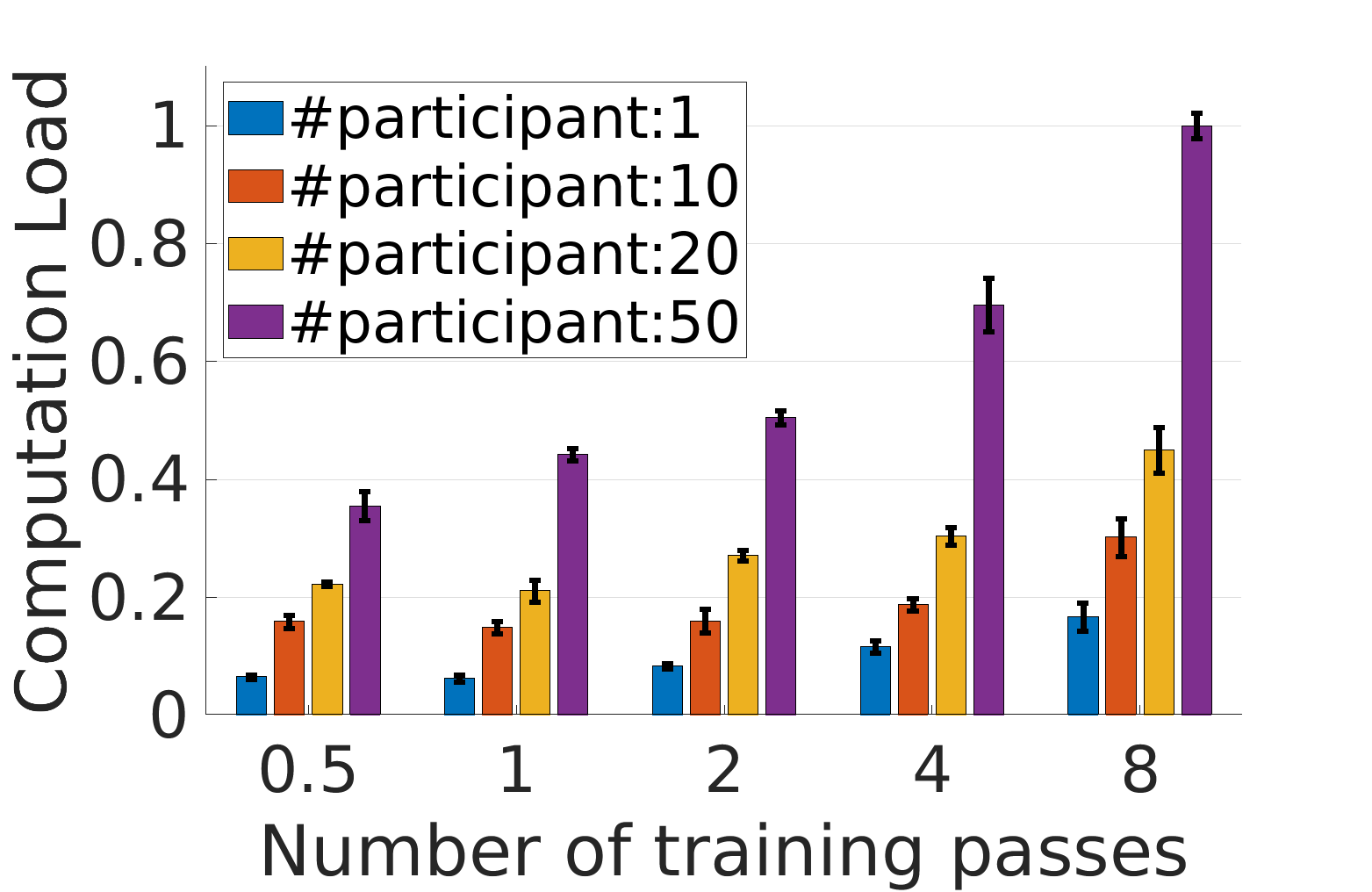}
    \label{fig:measure_M_E_compL}
}
\subfigure[Transmission Load]{
    \includegraphics[width=1.7in]{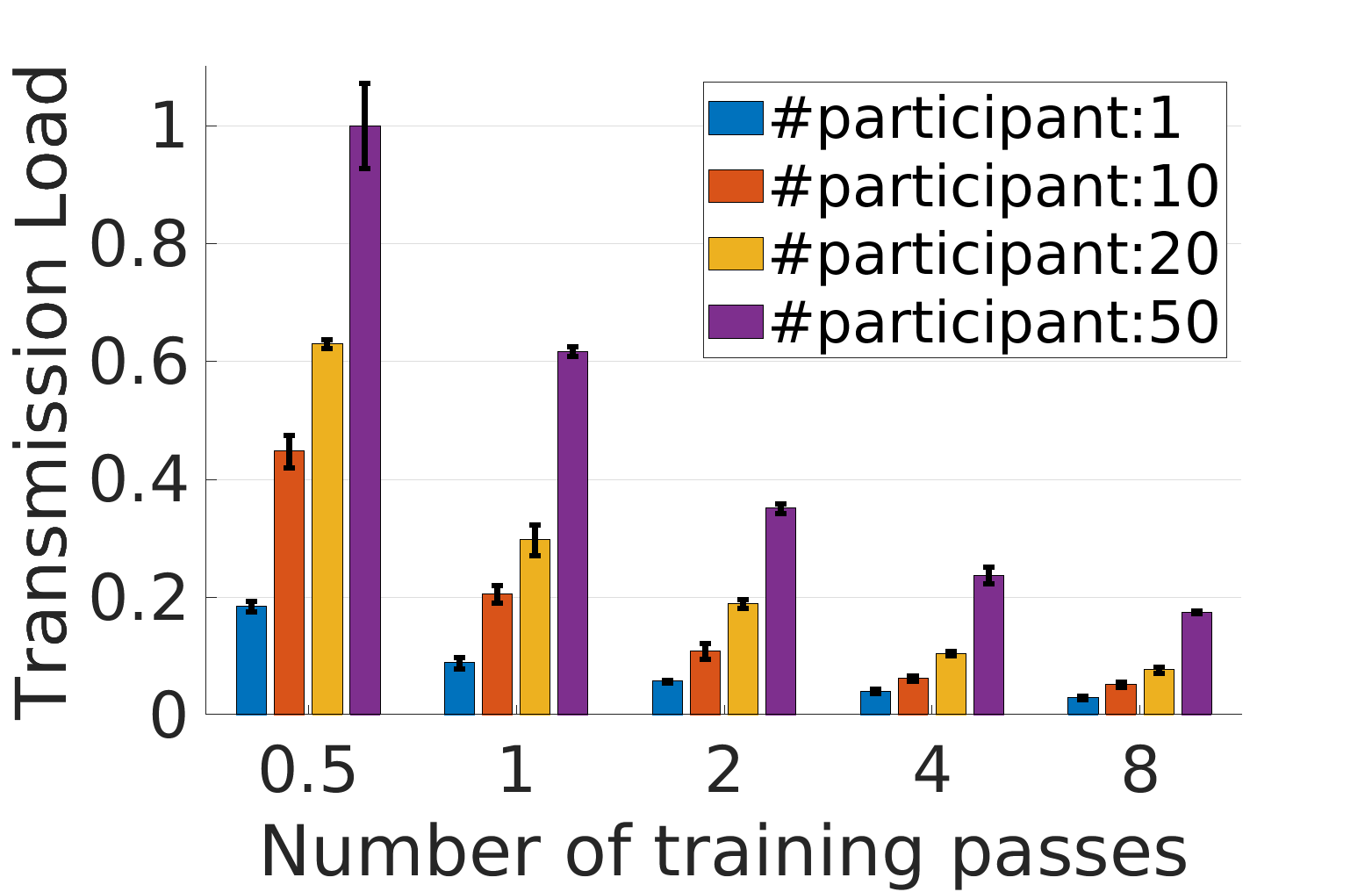}
    \label{fig:measure_M_E_transL}
}
}
    \caption{CompT, TransT, CompL, and TransL when a different number of participants and a different number of training passes are used. The lower the better.}
    \label{fig:measure_M_E}
\end{figure}

\noindent
\textbf{Computation Time (CompT)}.
Fig.~\ref{fig:measure_M_E_compT} compares CompT for a different number of participants $M$ and a different number of training passes $E$. In the experiments, we use ResNet-18 and normalize their overheads.  As we can see, more participants lead to smaller CompT, i.e., it takes a shorter time to converge. However, the difference is not significant among 10, 20, and 50 participants, especially when the number of training passes is large. 
In addition, we can see that larger $E$ has worse CompT. 

\noindent
\textbf{Transmission Time (TransT)}. Fig.~\ref{fig:measure_M_E_transT} plots TransT, which clearly shows that TransT favors larger $M$ and $E$. Since TransT is dependent on the number of training rounds $R$ (Eq.~(\ref{equ:transT})), it is equivalent to the metric of round-to-accuracy. Our measurement result is consistent with the common knowledge (e.g., \cite{fedNova20neurips}) that more participants and more training passes have a better round-to-accuracy performance. We can also observe that when $M$ is small, e.g., 1, TransL is much worse than the other cases.

\noindent
\textbf{Computation Load (CompL)}.
Fig.~\ref{fig:measure_M_E_compL} shows CompL. We make the following observations: (1) More participants result in worse CompL. The results indicate that the gain of faster model convergence from more participants does not compensate for the higher computation costs introduced by more participants. (2) CompL is increased when more training passes are used. This is probably because that larger $E$ diverges the model training~\cite{fedprox} and thus, the data utility per unit of computation cost is reduced.

\noindent
\textbf{Transmission Load (TransL)}.
Fig.~\ref{fig:measure_M_E_transL} illustrates TransL. As shown, more participants greatly increase TransL. This is because more participants can only weakly reduce the number of training rounds $R$~\cite{fedAvgConverge20iclr}, however, in each round, the number of transmissions is linearly increased with the number of participants. Regarding the number of training passes,  larger $E$ reduces the total number of training rounds $R$ and thus has better TransL. On the other hand, the gain of larger $E$ is diminishing. The results are consistent with the analysis of \cite{fedAvgConverge20iclr} that $R$ is hyperbolic with $E$ (the turning point happens around 100-1000 in their experiments).

\begin{figure}[!t]
    \centering
\centerline{
    \subfigure[Computation time and load]{
    \includegraphics[width=1.6in]{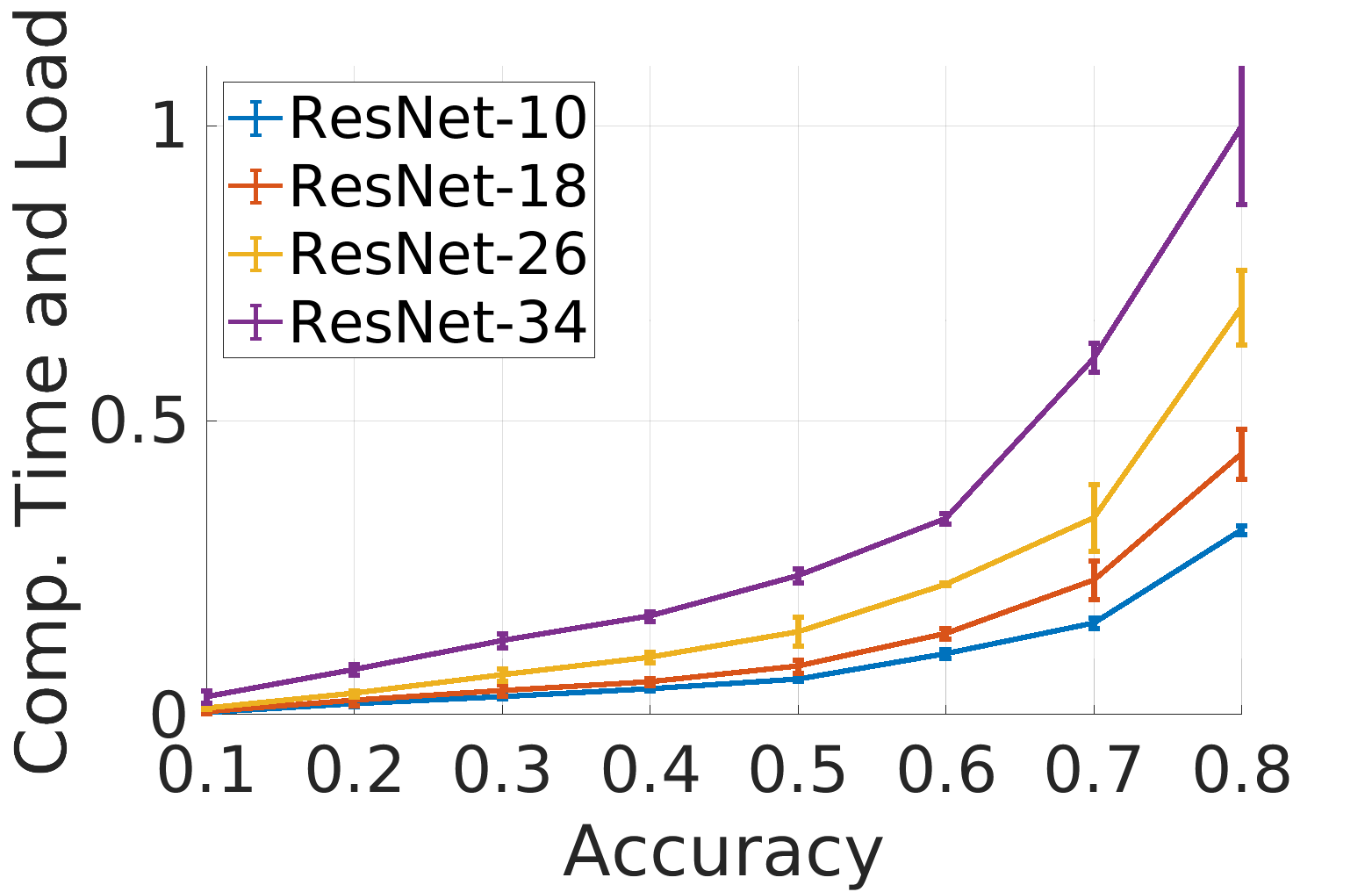}
    \label{fig:model_time_comp}
    }
    \subfigure[Transmission time and load]{
    \includegraphics[width=1.6in]{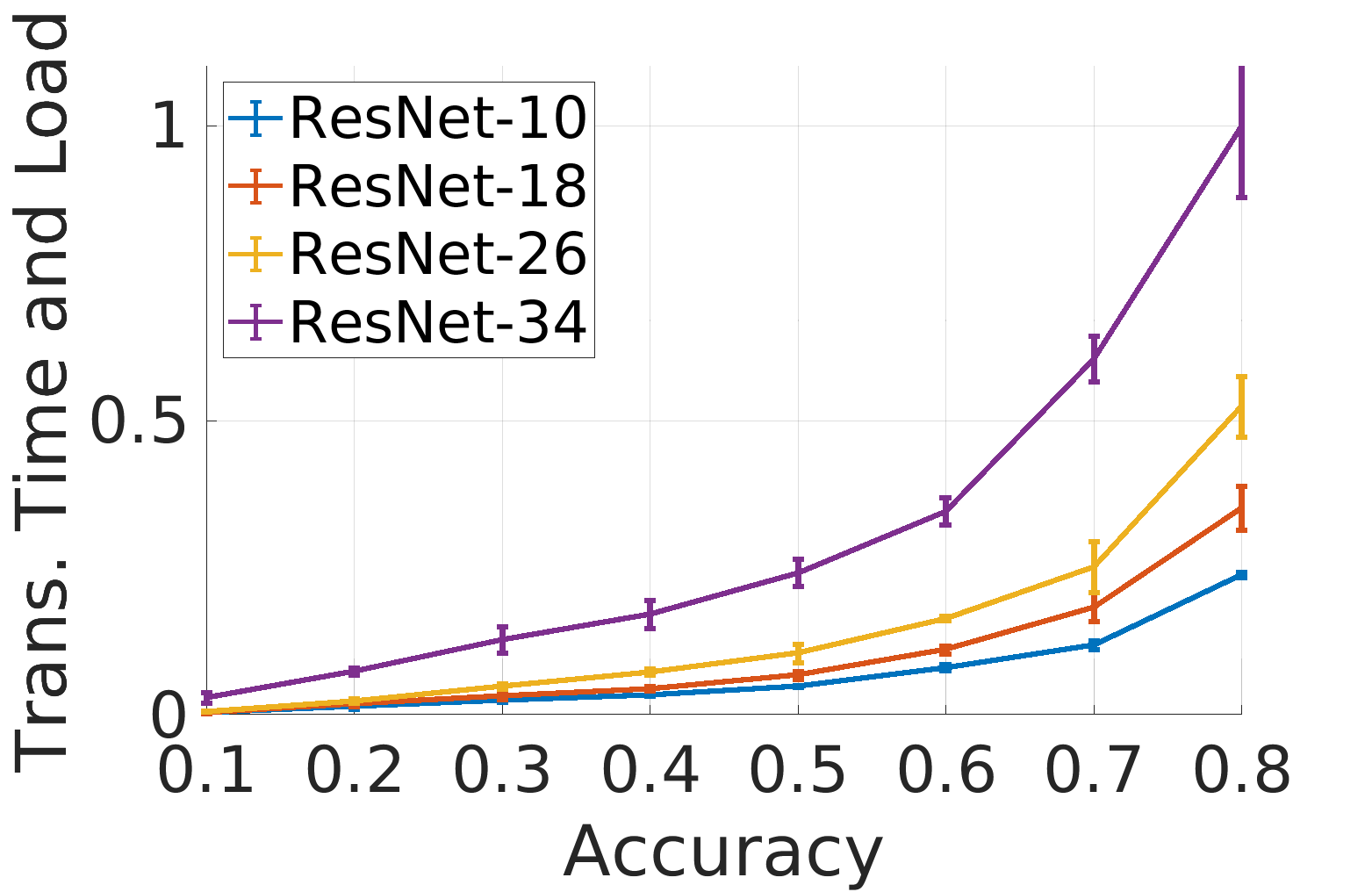}
    \label{fig:model_comm}
    }
}
    \caption{CompT, TransT, CompL, and TransL versus model complexity. The lower the better.}
    \label{fig:model}
\end{figure}

\noindent
\textbf{Model Complexity.}
Table \ref{tab:model} tabulates the models for comparing training overheads versus model complexity. In this experiment, we select one participant ($M=1$) to train one pass ($E=1$) on each training round. 
Fig.~\ref{fig:model} shows the normalized CompT, TransT, CompL, and TransL for different models. The x-axis is the target model accuracy, and the y-axis is the corresponding overhead to reach that model accuracy. Since only one client and one training pass are used on each round, CompT and CompL have the same normalized comparison, and so are TransT and TransL. The results show that smaller models are better with regard to all training aspects. 
In addition, it is interesting to note that heavier models have higher increase rates of overhead versus model accuracy. This means that model selection is especially essential for high model accuracy applications. 

\subsection{Summary of System Overheads}

Based on our measurement study, we summarize systems overheads versus FL hyper-parameters in Table \ref{tab:summary}.  As we can see, CompT, TransT, CompL, and TransL conflict with each other in terms of $M$ and $E$.
Regarding model complexity, smaller models have better system overhead if the model accuracy is satisfied. Please note that Table \ref{tab:summary} is consistent with existing works (e.g., \cite{fedNova20neurips}), but is more comprehensive. 

\section{FedTune}

\begin{table}[!t]
\centering
\centerline{
\scalebox{0.95}{
    \begin{tabular}{c | p{0.5in} p{0.5in} c}
      Training aspect & \centering $M$ & \centering $E$ & Model complexity \\\toprule
      CompT & \centering $>$ & \centering $<$ & $<$ \\ 
      TransT & \centering $>$ & \centering $>$ & $<$ \\ 
      CompL & \centering $<$ & \centering $<$ & $<$ \\
      TransL & \centering $<$ & \centering $>$ & $<$ \\
      Model Accuracy & \centering $=$ & \centering $=$ & $>$ \\\bottomrule
    \end{tabular}
}
}
\vspace{0.05in}
    \caption{System overheads versus the number of participants $M$, the number of training passes $E$, and model complexity. `$<$', `$=$', and `$>$' means the smaller the better, does not matter, and the larger the better, respectively.}
    \label{tab:summary}
\end{table}

\shortname considers training preferences for CompT, TransT, CompL, and TransL, denoted by $\alpha$, $\beta$, $\gamma$, and $\delta$, respectively. We have $\alpha + \beta + \gamma + \delta  = 1$. For example, $\alpha = 0.6$, $\beta = 0.2$, $\gamma = 0.1$, and $\delta = 0.1$ represent that the application is greatly concerned about CompT, while slightly about TransT, with CompL and TransL the least concern. 

\subsection{Problem Formulation}

For two sets of FL hyper-parameters $S_1$ and $S_2$, \shortname defines the comparison function $I(S_1, S_2)$ as
\begin{equation}
\begin{aligned}
    I(S_1, S_2) = \alpha \times \frac{t_2 - t_1}{t_1} + \beta \times \frac{q_2 - q_1}{q_1} \\ + \gamma \times \frac{z_2 - z_1}{z_1} + \delta \times \frac{v_2 - v_1}{v_1}
    \label{equ:compare}
\end{aligned}
\end{equation}
where 
$t_1$ and $t_2$ are CompT for $S_1$ and $S_2$ achieving the same model accuracy. Correspondingly, $q_1$ and $q_2$ are TransT, $z_1$ and $z_2$ are CompL, and $v_1$ and $v_2$ are TransL. If $I(S_1, S_2) < 0$, then $S_2$ is better than $S_1$.  A set of hyper-parameters is better than another set if the weighted improvement of some training aspects (e.g., CompT and CompL) is higher than the weighted degradation of the remaining training aspects (e.g., TransT and TransL). The weights are training preferences on CompT, TransT, CompL, and TransL.

However, the training overhead for different sets of FL hyper-parameters are unknown \textit{a priori}. As a result,  directly identifying the optimal hyper-parameters before FL training is impossible. Instead, we propose an iterative method to optimize the next set of hyper-parameters. Given the current set of hyper-parameters $S_{cur}$, the goal is to find a set of hyper-parameters $S_{nxt}$ that improves the training performance the most, that is, minimizing the following objective function:
\begin{equation}
\begin{aligned}
    G(S_{nxt}) =  \alpha \times \frac{t_{nxt} - t_{cur}}{t_{cur}} + \beta \times \frac{q_{nxt} - q_{cur}}{q_{cur}} \\ + \gamma \times \frac{z_{nxt} - z_{cur}}{z_{cur}}  + \delta \times \frac{v_{nxt} - v_{cur}}{v_{cur}}
\end{aligned}
\end{equation}
where $t_{cur}$, $q_{cur}$, $z_{cur}$, and $v_{cur}$ are CompT, TransT, CompL, and TransL under the current hyper-parameters $S_{cur}$;  $t_{nxt}$, $q_{nxt}$, $z_{nxt}$, and $v_{nxt}$ are CompT, TransT, CompL, and TransL for the next hyper-parameters $S_{nxt}$. We focus on the number of participants $M$ and the number of training passes $E$, since model complexity is monotonous with training overheads. Therefor, we need to optimize $S_{nxt} = \{M_{nxt}, E_{nxt}\}$. 

\subsection{$S_{nxt}$ Optimization}

To find the optimal $S_{nxt}$, we take the derivatives of $G(S_{nxt})$ over $M$ and $E$, obtaining
\begin{equation}
\begin{aligned}
    \Delta M = \frac{\partial G(S_{nxt})}{\partial M} =  \frac{\alpha}{t_{cur}} \times \frac{\partial t_{nxt}}{\partial M} + \frac{\beta}{q_{cur}} \times \frac{\partial q_{nxt}}{\partial M} \\ + \frac{\gamma}{z_{cur}} \times \frac{\partial z_{nxt}}{\partial M} + \frac{\delta}{v_{cur}} \times \frac{\partial v_{nxt}}{\partial M}
\end{aligned}
\end{equation}
\begin{equation}
\begin{aligned}
    \Delta E = \frac{\partial G(S_{nxt})}{\partial E} =  \frac{\alpha}{t_{cur}} \times \frac{\partial t_{nxt}}{\partial E} + \frac{\beta}{q_{cur}} \times \frac{\partial q_{nxt}}{\partial E} \\ + \frac{\gamma}{z_{cur}} \times \frac{\partial z_{nxt}}{\partial E} + \frac{\delta}{v_{cur}} \times \frac{\partial v_{nxt}}{\partial E}
\end{aligned}
\end{equation}

We illustrate how to obtain $\Delta M$. The process of solving $\Delta E$ is similar.
Considering that each step makes a small adjustment of $M$, $\partial t_{nxt} / \partial M$ can be represented by $(+1) \times |t_{nxt} - t_{cur}|$, where $(+1)$ means CompT prefers larger $M$ according to Table~\ref{tab:summary}. To estimate $| t_{nxt} - t_{cur} |$, we apply a linear function $\eta_{t-1} \times | t_{cur} - t_{prv}|$ where $\eta_{t-1} = \frac{|t_{cur} - t_{prv}|}{|t_{prv} - t_{prvprv}|}$ ($t_{prvprv}$ is the CompT at two steps before). Similarly, we have $\eta_{q-1}$,  $\eta_{z-1}$, $\eta_{v-1}$ for TransT, CompL, and TransL when calculating their derivatives over $M$.  
As a result, $\Delta M$ can be approximated as
\begin{equation}
\begin{aligned}
\Delta M = \frac{(+1) \times \alpha \times \eta_{t-1} \times |t_{cur} - t_{prv}| }{t_{cur}} \\ + \frac{(+1) \times \beta \times \eta_{q-1} \times |q_{cur} - q_{prv}| }{q_{cur}} \\ + \frac{(-1) \times \gamma \times \eta_{z-1} \times |z_{cur} - z_{prv}| }{z_{cur}} \\ + \frac{(-1) \times \delta \times \eta_{v-1} \times |v_{cur} - v_{prv}| }{v_{cur}}
\end{aligned}
\label{equ:Delta_M}
\end{equation}
Similarly, we can calculate $\Delta E$ as 
\begin{equation}
\begin{aligned}
\Delta E = \frac{(-1) \times \alpha \times \zeta_{t-1} \times |t_{cur} - t_{prv}| }{t_{cur}} \\ + \frac{(+1) \times \beta \times \zeta_{q-1} \times |q_{cur} - q_{prv}| }{q_{cur}} \\ + \frac{(-1) \times \gamma \times \zeta_{z-1} \times |z_{cur} - z_{prv}| }{z_{cur}} \\ + \frac{(+1) \times \delta \times \zeta_{v-1} \times |v_{cur} - v_{prv}| }{v_{cur}}
\end{aligned}
\label{equ:Delta_E}
\end{equation}
where $\zeta_{t-1}$, $\zeta_{q-1}$, $\zeta_{z-1}$, and $\zeta_{v-1}$ are the parameters for calculating the derivatives of CompT, TransT, CompL, and TransL over $E$. 


\subsection{Decision Making and Parameter Update}

\shortname is activated when the model accuracy is improved by at least $\epsilon$. 
Then, it computes $\Delta M$ and $\Delta E$, and determines the next $M$ and $E$ based on the signs of $\Delta M$ and $\Delta E$. Specifically, $M_{nxt} = M_{cur} + 1$ if $\Delta M > 0$, otherwise, $M_{nxt} = M_{cur} - 1$. Likewise, \shortname increases $E_{nxt}$ by one if $\Delta E > 0$; else \shortname decreases $E_{nxt}$ by one.  The FL training is resumed using the new hyper-parameters.  \shortname is lightweight and negligible to the FL training: it only requires dozens of multiplication and addition calculations.

\shortname automatically updates $\eta_{t-1}$, $\eta_{q-1}$,  $\eta_{z-1}$, $\eta_{v-1}$, $\zeta_{t-1}$, $\zeta_{q-1}$,  $\zeta_{z-1}$, and $\zeta_{v-1}$ during FL training. At each step, \shortname updates the parameters that favor the current decision. For example, if $M_{cur}$ is larger than $M_{prv}$, \shortname updates $\eta_{t-1}$ and $\eta_{q-1}$ as CompT and TransT prefer larger $M$; otherwise, \shortname updates $\eta_{z-1}$ and $\eta_{v-1}$. 

Furthermore, \shortname incorporates a penalty mechanism to mitigate bad decisions. Given the previous hyper-parameters $S_{prv}$ and the current hyper-parameters $S_{cur}$, \shortname calculates the comparison function $I(S_{prv}, S_{cur})$. A bad decision occurs if the sign of $I(S_{prv}, S_{cur})$ is positive. In this case, \shortname multiplies the parameters that are against the current decision by a constant penalty factor, denoted by $D$ ($D \geq 1$). For example, if $I(S_{prv}, S_{cur}) > 0$ and $M_{cur} > M_{prv}$, \shortname updates $\eta_{t-1}$ and $\eta_{q-1}$ as explained before, but also multiplies $\eta_{z-1}$ and $\eta_{v-1}$ by $D$.

\section{Experiments and Analysis}

\noindent
\textbf{Benchmarks and Baseline}. 
We evaluate \shortname on three datasets: speech-to-command~\cite{speechToCommandDataset}, EMNIST~\cite{emnist}, and Cifar-100~\cite{cifar100}, and three aggregation methods: FedAvg~\cite{FedAvg17aistats}, FedNova~\cite{fedNova20neurips}, and FedAdagrad~\cite{fedyogi}.  We set equal values for the combination of training preferences $\alpha$, $\beta$, $\gamma$ and $\delta$ (see the first column in Table~\ref{tab:eva_speech_to_command}). Therefore, for each dataset, we conduct 15 combinations of training preferences. We set target model accuracy for each dataset and measure CompT, TransT, CompL, and TransL for reaching the target model accuracy. We regard the practice of using fixed $M$ and $E$ as the baseline and compare \shortname to the baseline by calculating Eq.~(\ref{equ:compare}).
We implemented \shortname in PyTorch. All the experiments are conducted on a server with 24-GB Nvidia RTX A5000 GPUs.


\subsection{Overall Performance}
\label{sect:eva_overall}

\begin{table}[!t]
    \centering
\scalebox{0.9}{
    \centerline{
    \begin{tabular}{c |  c c  c}
\toprule
Dataset & Speech-command & EMNIST & Cifar-100 \\
Data Feature & Voice & Handwriting & Image \\ 
ML Model & ResNet-10 & 2-layer MLP & ResNet-10 \\[3pt]\hline 
\\[-5pt]
Performance & +22.48\% (17.97\%) & +8.48\% (5.51\%) & +9.33\% (5.47\%) \\
 \bottomrule
    \end{tabular}
    }
    }
\vspace{0.05in}
    \caption{Performance of \shortname for diverse datasets when FedAvg aggregation method is applied. }
    \label{tab:all_dataset}
\end{table}

\noindent
\textbf{Training setup}. (1) \textit{speech-to-command} dataset. It classifies audio clips to 35 commands (e.g., `yes', `off'). We transform audio clips to 64-by-64 spectrograms and then downsize them to 32-by-32 gray-scale images. As officially suggested~\cite{speechToCommandDataset}, we use 2112 clients' data for training and the remaining 506 clients' data for testing. We set the mini-batch size to 5, considering that many clients have few data points. We use ResNet-10 and the target model accuracy of 0.8. (2) \textit{EMNIST} dataset. It classifies handwriting (28-by-28 gray-scale images) into 62 digits and letters (lowercase and uppercase). We split the dataset based on the writer ID. We randomly select 70\% writers' data for training and the remaining for testing. We use a Multiplayer Perception (MLP) model with one hidden layer~(200 neurons with ReLu activation). We set the mini-batch size to 10 and the target model accuracy of 0.7. (3) \textit{Cifar-100} dataset. It classifies 32-by-32 RGB images to 100 classes. We randomly split the dataset into 1200 users, where each user has 50 data points. Then, we randomly select 1000 users for training and the remaining 200 users for testing. We set the mini-batch size to 10. ResNet-18 is used, and the target model accuracy is set to 0.2 (due to our limited computational capability, we set a low threshold for Cifar-100).

For all datasets, we normalize the input images with the mean and the standard deviation of the training data before feeding them to models for training and testing. Both $M$ and $E$ are initially set to 20. \shortname is activated when the model accuracy is increased by at least 0.01 (i.e., $\epsilon = 0.01$). The penalty factor $D$ is set to 10. All results are averaged by three experiments.


\begin{table}[!t]
    \centering
\scalebox{0.9}{
    \centerline{
    \begin{tabular}{c | c c c}
    \toprule
Aggregator & FedAvg & FedNova & FedAdagrad \\
Performance & +22.48\% (17.97\%) & +23.53\% (6.64\%) & +26.75\% (6.10\%) \\
\bottomrule
    \end{tabular}
    }
    }
    \vspace{0.05in}
    \caption{Performance of \shortname for diverse aggregation algorithms. Speech-to-command dataset and ResNet-10 are used in this experiment.}
    \label{tab:all_aggregator}
\end{table}


\begin{table*}[h]
    \centering
    \centerline{
\scalebox{0.92}{
\begin{tabular}{c c c c | c c c c | c c | c}
$\alpha$ & $\beta$ & $\gamma$  & $\delta$ & CompT ($10^{12}$) & TransT ($10^6$) & CompL ($10^{12}$) & TransL ($10^6$) & Final M & Final E & Overall \\\toprule 
- & - & - & - & 0.94 (0.01) & 11.61 (0.10) & 5.97 (0.04) & 232.24 (1.99) & 20 & 20 & - \\
1.0 & 0.0 & 0.0 & 0.0 & 0.42 (0.02) & 50.19 (2.57) & 4.57 (0.22) & 2418.71 (240.91) & 57.33 (4.50) & 1.00 (0.00) & +55.23\% (2.22\%) \\
0.0 & 1.0 & 0.0 & 0.0 & 1.34 (0.22) & 7.68 (1.12) & 14.99 (2.73) & 289.82 (46.98) & 48.00 (2.16) & 48.00 (2.16) & +33.87\% (9.67\%) \\
0.0 & 0.0 & 1.0 & 0.0 & 1.02 (0.10) & 615.98 (97.52) & 1.76 (0.16) & 672.21 (91.62) & 1.00 (0.00) & 1.00 (0.00) & +70.51\% (2.75\%) \\
0.0 & 0.0 & 0.0 & 1.0 & 2.18 (0.47) & 35.47 (7.51) & 3.30 (0.22) & 76.47 (1.68) & 1.00 (0.00) & 46.67 (3.30) & +67.07\% (0.72\%) \\
0.5 & 0.5 & 0.0 & 0.0 & 0.82 (0.13) & 9.17 (1.26) & 9.13 (1.66) & 347.11 (54.31) & 47.33 (2.05) & 21.33 (4.78) & +16.97\% (9.68\%) \\
0.5 & 0.0 & 0.5 & 0.0 & 0.48 (0.04) & 81.42 (9.83) & 3.23 (0.14) & 1875.99 (155.21) & 25.00 (1.63) & 1.00 (0.00) & +47.57\% (3.43\%) \\
0.5 & 0.0 & 0.0 & 0.5 & 0.79 (0.10) & 11.59 (0.55) & 5.04 (0.89) & 241.86 (68.65) & 22.33 (5.79) & 15.67 (4.50) & +5.82\% (11.28\%) \\
0.0 & 0.5 & 0.5 & 0.0 & 0.83 (0.03) & 10.66 (0.15) & 5.16 (0.31) & 207.79 (6.08) & 21.00 (1.41) & 21.00 (1.41) & +10.87\% (2.83\%) \\
0.0 & 0.5 & 0.0 & 0.5 & 1.54 (0.16) & 11.48 (3.83) & 9.59 (3.52) & 190.52 (61.53) & 19.67 (14.82) & 49.00 (0.00) & +9.55\% (7.08\%) \\
0.0 & 0.0 & 0.5 & 0.5 & 1.69 (0.26) & 50.14 (8.21) & 2.70 (0.26) & 93.21 (8.48) & 1.00 (0.00) & 23.33 (2.49) & +57.32\% (3.76\%) \\
0.33 & 0.33 & 0.33 & 0.0 & 0.82 (0.07) & 11.59 (1.01) & 5.65 (0.27) & 255.35 (9.65) & 22.33 (2.62) & 15.67 (1.25) & +6.09\% (6.67\%) \\
0.33 & 0.33 & 0.0 & 0.33 & 1.06 (0.08) & 10.07 (0.90) & 8.10 (0.34) & 247.54 (29.18) & 26.33 (2.05) & 27.00 (2.16) & -1.93\% (7.40\%) \\
0.33 & 0.0 & 0.33 & 0.33 & 0.91 (0.19) & 18.23 (5.83) & 4.15 (1.13) & 229.26 (63.40) & 12.00 (1.41) & 14.00 (5.72) & +11.66\% (11.76\%) \\
0.0 & 0.33 & 0.33 & 0.33 & 1.13 (0.13) & 16.16 (3.36) & 4.51 (0.59) & 169.93 (25.84) & 9.00 (5.35) & 23.00 (4.55) & +3.99\% (6.19\%) \\
0.25 & 0.25 & 0.25 & 0.25 & 0.91 (0.10) & 9.73 (1.81) & 6.19 (0.76) & 207.34 (3.34) & 23.33 (5.44) & 22.67 (3.30) & +6.51\% (6.13\%) \\\bottomrule
\end{tabular}
}
    }
    \vspace{0.05in}
    \caption{Performance of \shortname for the speech-to-command dataset when FedAdagrad is used for aggregation. \\`$+$' is improvement and `$-$' is degradation. Standard deviation in parentheses. }
    \label{tab:eva_speech_to_command}
    \vspace{-0.25in}
\end{table*}

\noindent
\textbf{Results for Diverse Datasets}. 
Table \ref{tab:all_dataset} shows the overall performance of \shortname for different datasets when FedAvg is applied. We set the learning rate to 0.01 for the speech-to-command dataset and the EMNIST dataset, and 0.1 for the Cifar-100 dataset, all with the momentum of 0.9. We show the standard deviation in parenthesis. As shown, \shortname consistently improves the system performance across all the three datasets. In particular, \shortname reduces 22.48\% system overhead of the speech-to-command dataset compared to the baseline. 
We also observe that the FL training benefits more from \shortname if the training process needs more training rounds to converge. Our experiments with EMNIST~(small model) and Cifar100 (low target accuracy) only require a few dozens of training rounds to reach their target model accuracy, and thus their performance gains from \shortname are not significant. The observation is consistent with the decision-making process in \shortname, which increases/decreases hyper-parameters by only one at each step. We leave it as future work to augment \shortname to change hyper-parameters with adaptive degrees.  

\noindent
\textbf{Results for Different Aggregation Methods}. Table \ref{tab:all_aggregator} shows the overall performance of \shortname for different aggregation methods when we use the speech-to-command dataset and the ResNet-10 model. We set the learning rate to 0.1, $\beta_1$ to 0, and $\tau$ to 1e-3 in FedAdagrad. As shown, \shortname achieves consistent performance gain for diverse aggregation methods. In particular, FedAdagrad reduces the system overhead by 26.75\%.  





\noindent
\textbf{Trace Analysis of \shortname}. We present the details of traces when the speech-to-command dataset and the FedAdagrad aggregation method are used. 
Table \ref{tab:eva_speech_to_command} tabulates the results. We report the average performance, as well as their standard deviations in parentheses. The first row is the baseline, which does not change hyper-parameters during the FL training. 
We show the final $M$ and $E$ when the training is finished.  As we can see from Table \ref{tab:eva_speech_to_command}, \shortname can adapt to different training preferences. Only one preference (0.33, 0.33, 0, 0.33) results in a slightly degraded performance. On average, \shortname improves the overall performance by 26.75\%. 

\section{Conclusion}


FL involves high system overheads, which hinders its research and real-world deployment. We argue that optimizing system overhead for FL applications is valuable. To this end, in this work, we propose \shortname to adjust FL hyper-parameters, catering to the application's training preferences automatically. 
Our evaluation results show that \shortname is general, lightweight,  flexible, and is able to significantly reduce system overhead. 

\ifCLASSOPTIONcompsoc
  \section*{Acknowledgments}
\else
  \section*{Acknowledgment}
\fi

This research was partially sponsored by the U.S. Army Combat Capabilities Development Command Army Research Laboratory and was accomplished under Cooperative Agreement Number W911NF-13-2-0045 (ARL Cyber Security CRA). The views and conclusions contained in this document are those of the authors and should not be interpreted as representing the official policies, either expressed or implied, of the Combat Capabilities Development Command Army Research Laboratory or the U.S. Government. The U.S. Government is authorized to reproduce and distribute reprints for Government purposes notwithstanding any copyright notation here on. The research was also partially supported by NSF through CNS 1901218 and USDA-020-67021-32855.

\bibliographystyle{abbrv}
\bibliography{iclr2022_conference}

\end{document}